\documentclass[sigconf]{acmart}

\AtBeginDocument{%
  }

\setcopyright{acmlicensed}
\copyrightyear{2024}
\acmYear{2024}
\acmDOI{XXXXXXX.XXXXXXX}

\acmConference[CMLKDD2024, KDD 2024 Workshop]{Causal Inference and Machine Learning in Practice}{August 2024}{Barcelona, Spain}
\acmBooktitle{CMLKDD2024: KDD 2024 Workshop,
 Aug 25--29, 2024, Barcelona, Spain}
\acmISBN{978-1-4503-XXXX-X/18/06}

\newtheorem{ass}{\bf Assumption}

\newcommand{\indep}{\perp \!\!\! \perp}
\newcommand{\notindep}{\not \! \perp \!\!\! \perp}

\usepackage{algorithmic}
\usepackage[ruled]{algorithm2e}
\begin{document}

\title[A Partial Initialization Strategy for Treatment Effect Estimation with Hidden Confounding]{A Partial Initialization Strategy to Mitigate the Overfitting Problem in CATE Estimation with Hidden Confounding}

\author{Chuan Zhou}
\authornote{Both authors contributed equally to this research.}
\email{zhouchuancn@pku.edu.cn}
\affiliation{%
  \institution{Peking University \& MBZUAI}
  \city{Beijing}
  \country{China}
}
\author{Yaxuan Li}
\authornotemark[1]
\authornote{This work was done during the research internship of Yaxuan Li at Peking University.}
\email{yaxuanli.cn@gmail.com}
\affiliation{%
  \institution{Peking University}
  \city{Beijing}
  \country{China}
}

\author{Chunyuan Zheng}
\email{cyzheng@stu.pku.edu.cn}
\affiliation{%
  \institution{Peking University \& MBZUAI}
  \city{Beijing}
  \country{China}}

\author{Haiteng Zhang}
\email{zhanghaiteng22@mails.ucas.ac.cn}
\affiliation{%
 \institution{Chinese Academy of Sciences \& MBZUAI}
  \city{Beijing}
  \country{China}
}

\author{Haoxuan Li}
\authornote{Haoxuan Li and Mingming Gong are corresponding authors.}
\email{hxli@stu.pku.edu.cn}
\affiliation{%
 \institution{Peking University \& MBZUAI}
 \city{Beijing}
 \country{China}}

\author{Mingming Gong}
\authornotemark[3]
\email{mingming.gong@unimelb.edu.au}
\affiliation{%
  \institution{The University of Melbourne \& MBZUAI}
  \city{Melbourne}
  \country{Australia}}
\renewcommand{\shortauthors}{Chuan Zhou et al.}

\begin{abstract}
Estimating the conditional average treatment effect (CATE) from observational data plays a crucial role in areas such as e-commerce, healthcare, and economics. Existing studies mainly rely on the strong ignorability assumption that there are no hidden confounders, whose existence cannot be tested from observational data and can invalidate any causal conclusion. In contrast, data collected from randomized controlled trials (RCT) do not suffer from confounding but are usually limited by a small sample size. To avoid overfitting caused by the small-scale RCT data, we propose a novel two-stage pretraining-finetuning (TSPF) framework with a partial parameter initialization strategy to estimate the CATE in the presence of hidden confounding. In the first stage, a foundational representation of covariates is trained to estimate counterfactual outcomes through large-scale observational data. In the second stage, we propose to train an augmented representation of the covariates, which is concatenated with the foundational representation obtained in the first stage to adjust for the hidden confounding. Rather than training a separate network from scratch, part of the prediction heads are initialized from the first stage. The superiority of our approach is validated on two datasets with extensive experiments.
\end{abstract}




\maketitle

\section{Introduction}


The conditional average treatment effect (CATE) is the average causal effect of a treatment or an intervention on the outcome of interest given the covariates \cite{pearl2009causality}, which plays an important role in diverse fields, such as trustworthy artificial intelligence~\cite{li2023trustworthy}, electronic commerce \cite{ijcai2022p787}, healthcare \cite{robins2016causal}, and economics \cite{huynh2016causal}. In e-commerce, the platforms desire to predict how recommending a specific product to a particular user affects the probability of purchase~\cite{li2022recommender}, and thereby influence the total profit. In healthcare, doctors assess the potential outcome for different patient groups when receiving a certain treatment~\cite{corrao2012external} for precision medicine. Similarly in economics, the policymakers evaluate how much a job training program will raise employment opportunities for certain groups of unemployed individuals \cite{britto2022effect}.

To enhance the accuracy of CATE estimation, representation-based learning approaches have gathered increasing attention due to their impressive performance \cite{johansson2016learning,lilearning,zhu2024contrastive,wang2024optimal}. These approaches focus on generating covariate representations, to mitigate confounding bias by minimizing distributional discrepancies between the treatment and control groups. Previous approaches have developed substantial theory and explored extensive practice to obtain such representations. For instance, some of them use integral probability metric (IPM) for regularization \cite{johansson2016learning}, while a few approaches emphasize local similarity preservation \cite{yao2018representation}, targeted learning \cite{zhang2020learning}, or view the problem from new perspectives like optimal transport \cite{wang2024optimal} and out-of-distribution~\cite{ligenerative}.

However, the aforementioned methods may ignore hidden confounding, which is very common in real-world scenarios~\cite{li2024removing}. In our e-commerce example, the financial status of users might be sensitive and difficult to collect \cite{li2023balancing}. In the healthcare case, the personal lifestyles of patients are difficult to obtain \cite{charpignon2022causal}. For the example in economics, personal working status is difficult to measure~\cite{xu2020relationship}. These hidden variables can affect treatment and outcome simultaneously, which causes confounding bias in the estimation of causal effects. Therefore, proposing methods to account for the confounding bias is crucial to accurately estimate CATE.

To address the hidden confounding problem, one category of mainstream methods can rely only on large-scale observational (OBS) data, including sensitivity analysis \cite{imbens2003sensitivity, dorn2024doubly}, front-door adjustment methods \cite{fulcher2020robust}, and instrumental variables methods \cite{Angrist1996}. These methods require additional strong assumptions that cannot be tested from the data and raise concerns if these assumptions are violated \cite{hartwig2023average, kongtowards}. Compared to the OBS data, randomized controlled trial (RCT) data are considered as the gold standard for causal effect estimation \cite{prosperi2020causal}. However, practical challenges such as high costs and ethical concerns may make the collection of RCT data difficult \cite{zabor2020randomized, bedecarrats2020randomized}, resulting in limited sample sizes. Due to the small size, it is impractical to directly train causal effect prediction models on RCT data alone \cite{hoogland2021tutorial}. Therefore, it is necessary to find an effective method to combine small-scale RCT data with large-scale OBS data. However, previous methods either rely on untestable strong assumptions~\cite{, kallus2018removing, hatt2022generalizing} or suffer from the risk of overfitting due to limited RCT sample sizes~\cite{colnet2024causal,wu2022integrative}.

In this paper, we introduce a two-stage pretraining-finetuning (TSPF) framework with a partial initialization strategy for CATE estimation under hidden confounding. Our approach leverages large-scale OBS data to train a foundational representation of covariates and then uses relatively small-scale RCT data to adjust the bias in the representation learned from OBS data. We then train an unbiased prediction model using this adjusted representation. In the second stage, we introduce an additional module that ensures stronger representation ability compared to the methods that use RCT data to estimate the residuals and initialize part of the prediction heads with parameters from the first stage.

The contributions of this paper are summarized as follows.
\begin{itemize}
    \item We present the TSPF framework that includes a partial initialization strategy for CATE estimation, tackling hidden confounding by using a small amount of unconfounded RCT data to adjust the representations learned from OBS data.
    \item The proposed framework does not rely on the linear and additive generation assumptions, and can flexibly adjust its model structure according to the sample size of RCT data, thus mitigating the over-fitting problem. 
    \item Experiments conducted on the semi-synthetic IHDP and Jobs datasets demonstrate the effectiveness of our approach.
\end{itemize}

\section{Preliminaries}
\subsection{Problem Setup}
We consider two independent data sources taken from the same target population: one from OBS and the other from RCT. Each individual in the OBS or RCT study is an observation of $(X, Y, T, G)$, a random tuple with distribution $P$. For the $i$-th individual, the observation comprises $d$-dimensional covariates $X_i \in \mathcal{X} \subseteq \mathbb{R}^d$, the observed outcome $Y_i$, the assigned binary treatments $T_i \in \{0,1\}$ ($T_i = 0$ for the controlled and $T_i = 1$ for the treated individuals) and $ G_i $ denoting participation in the OBS ($ G_i = 0 $) or RCT ($G_i= 1$) study. Using the Neyman-Rubin potential outcome framework \cite{imbens2015causal}, we let $Y^1_i, Y^0_i$ be the potential outcomes. We denote the OBS data as $\mathcal{D}^{OBS}=\{(X_i, T_i, Y_i, G_{i}=0):i \in \mathcal{O}\}$ with sample size $n$, and the RCT data as $\mathcal{D}^{RCT}=\{(X_i, T_i, Y_i, G_{i}=1):i \in \mathcal{R}\}$ with sample size $m$, where $\mathcal{O}=\{1,\ldots, n\}$ and $\mathcal{R}=\{n+1,\ldots,n+m\}$ are sample index sets for the OBS and RCT data, respectively. The total sample size is $N = n + m$.  We define the propensity score as $e(x,G)=P(T=1\mid X=x,G)$. The CATE is defined as the conditional expectation of difference between potential outcomes under the treatment group and the control group as follows:
\begin{equation*}
    \tau(x)=\mathbb{E}[Y^1-Y^0\mid X=x].
\end{equation*}

\subsection{Identification of CATE}
To identify the CATE from observed data, in addition to the Stable Unit Treatment Value Assumption (SUTVA) that there are no interference between units and there are no different forms of each treatment level, the following three assumptions are required:
\begin{ass}
\label{ass1}(Ignorability)
$(Y^1, Y^0)\indep T\mid X$,
\end{ass}
\begin{ass}
\label{ass2}
(Consistency) $Y=TY^1 +(1-T)Y^0$,
\end{ass}
\begin{ass}
\label{ass3}
(Positivity) $0<e(x,G=1)<1$.
\end{ass}
Assumption \ref{ass1} is also known as no hidden confounding, which holds in the RCT by default due to the randomized treatment assignment. We can identify CATE based on the RCT data:
\begin{equation*}
    \tau(x)=\mathbb{E}[Y\mid T=1,X=x,G=1]-\mathbb{E}[Y\mid T=0,X=x,G=1].
\end{equation*}

The unconfoundedness assumption is not assumed to hold for the observational data, i.e. $(Y^1,Y^0)\notindep T\mid (X,G=0)$. We cannot identify $\tau(x)$ based on OBS data only. Let us denote the difference in conditional average outcomes in the observational data by:
\begin{equation*}
    \zeta(x)=\mathbb{E}\left[Y\mid T=1,X=x,G=0\right]-\mathbb{E}\left[Y\mid T=0,X=x,G=0\right].
\end{equation*}

Note that due to hidden confounding, $\zeta(x)\neq \tau(x)$ for any $x$. The difference between these two quantities is precisely the confounding effect, which we denote the residual function as:
\begin{equation*}
    \eta(x)=\tau(x)-\zeta(x).
\end{equation*}

\section{Methodology}
\begin{figure*}
    \centering
    \includegraphics[width=1.0\linewidth]{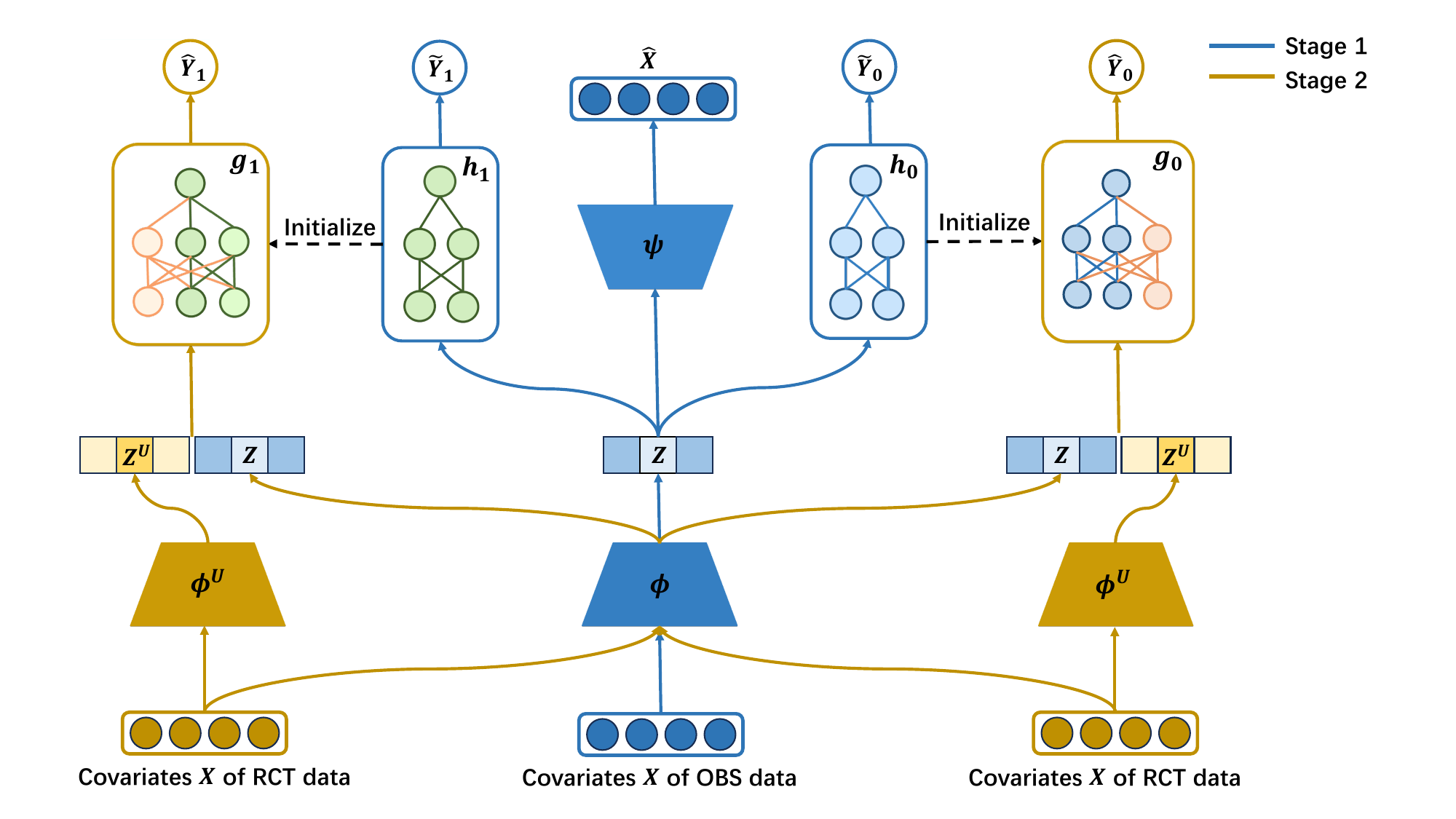}
    \caption{The framework of our proposed method, composed of the modules for the first stage (blue) and second stage (yellow). Note that the two $\phi^U$ shown in the figure represent the same module.}
    \label{fig1}
\end{figure*}
In this section, we present a two-stage framework for the estimation of CATE based on the pretraining-finetuning principle, as shown in Figure \ref{fig1}. The motivation is to use large-scale OBS data to train a foundational representation of covariates, then to use relatively small-scale unbiased RCT data to remove the bias in the representation learned from OBS data for training the unbiased prediction model. Specifically, in the first stage, only the OBS data is used. We start with a representation module $\phi$, followed by one reconstruction module $\psi$ and two prediction heads $h_0$ and $h_1$ to ensure the learned covariate representation can have enough information and predict the outcome for control group and treatment group simultaneously. While in the second stage, we use only the RCT data. The representation module $\phi$ learned in the first stage is frozen, with a learnable representation adapter module $\phi^{U}$ to remove the bias of $\phi$. Then the concatenated representation is fed to two prediction heads $g_0$ and $g_1$ to obtain the unbiased predicted potential outcomes under control and treatment groups respectively. We carefully design an initialization strategy to ensure that the initialized second-stage model produces the same predictions as the converged model in the first stage. Our approach distinguishes from the one proposed by Kallus et al.~\cite{kallus2018removing}, which only uses linear regression to estimate the residual function $\eta(x)$ in the second stage. We can regard the first stage as pretraining on the large OBS data and the second stage as finetuning on the small-scale unbiased RCT data.

\subsection{First Stage: Pretraining Stage}
The goal of our first-stage training is to obtain a representation module as well as prediction heads that can accurately estimate the potential outcomes of OBS data. These modules offer high-quality initialization for second-stage training, allowing fine-tuning on the RCT data to avoid the overfitting problem. 
A three-headed architecture and a multi-task training framework are employed to achieve this goal. Next, we will look into the details of each module.
\subsubsection{Representation}
We design a multi-layer feed-forward neural network $\phi$ to obtain a representation $Z$ for the covariates $X$ for both treatment and control groups. In other words, for an individual sample $(X_i,T_i,Y_i,G_i=0)$, the representation $Z_i=\phi(X_i)$ remains the same whether $T_i=0$ or $T_i=1$. To better estimate the causal effect, we adopt the idea of covariate balancing on the representation space as Shalit et al.~\cite{shalit2017estimating}. Specifically, the representations for treatment group $\{Z_i=\phi(X_i):G_i=0, T_i=1\}$ are regarded as i.i.d samples drawn randomly from a distribution $P_{\phi}^{t=1}$ and similarly $P_{\phi}^{t=0}$ for the control group. We anticipate the distributions of representations to be similar between the treatment and control groups. An integral probability metric (IPM) is employed to measure the distance between the two distributions. Thus the covariate unbalancing loss is defined as:
\begin{equation*}
\mathcal{L}_{unb}=\mathrm{IPM}_{\mathcal{G}}(\hat{P}_{\phi}^{t=1}, \hat{P}_{\phi}^{t=0}),
\end{equation*}
where $\mathrm{IPM}_{\mathcal{G}}(\cdot)$ is the empirical IPM defined by the function family $\mathcal{G}$, and $\hat{P}_{\phi}^{t=1}$ and $\hat{P}_{\phi}^{t=0}$ are empirical distributions of $P_{\phi}^{t=1}$ and $P_{\phi}^{t=0}$ respectively. In the implementation, we adopt the Wasserstein distance as a showcase, which can be consistently estimated from finite samples within a mini-batch~\cite{frogner2015learning}.
\subsubsection{Reconstruction and Prediction}
To ensure that $Z$ retains as much information about the original covariates as possible, we introduce the decoder network $\psi$ to reconstruct the original covariates: $\hat{X}=\psi(Z)=\psi(\phi(X))$. The reconstruction loss is computed by the mean squared error (MSE):
\begin{equation*}
\mathcal{L}_{rec}=\dfrac{1}{|\mathcal{O}|}\sum_{i\in \mathcal{O}} ||\hat{X}_i- X_i||_2^2,
\end{equation*}
where $\hat{X}_{i}=\psi(\phi(X_i))$ is reconstructed covariate for the $i$-th sample in the OBS data. The reconstruction design resembling an autoencoder allows the learned representations to encompass nearly complete information in the covariates, rather than only the information necessary for fitting the training set, thereby enhancing the generalization of our representation module.

We then use the representations $Z$ to estimate the potential outcomes with two $l_p-$layer prediction heads $h_0$ and $h_1$, which are the predictors for control and treatment outcomes, respectively. Note that hidden confounding in the observational data can lead to biased estimations of potential outcomes, we refer to the prediction result $\tilde{Y}^0=h_0(Z)=h_0(\phi(X))$ as the pseudo control outcome and $\tilde{Y}^1=h_1(Z)=h_1(\phi(X))$ as the pseudo treatment outcome. To enhance comparability between the treatment group and control group, we employ a reweighting technique to balance the two groups. Formally, let $f_h(x,t)=h_t (\phi(x))$ with $t\in \{0,1\}$ be the predicted potential outcomes by via the two heads $h_0$ and $h_1$, the loss for outcome prediction is as follows:
\begin{equation*}
 \mathcal{L}_f=\frac{1}{|\mathcal{O}|}\sum_{i\in \mathcal{O}}w_i\cdot l(Y_{i},f_h(X_i, T_i)),
\end{equation*}
with $w_i=\frac{T_i}{2u}+\frac{1-T_i}{2(1-u)}$, where $u=\frac{1}{n}\sum_{i=1}^{n}T_i$. The loss function $l(\cdot, \cdot)$ in $\mathcal{L}_f$ is flexible and can be determined based on the value range of potential outcomes. If the potential outcomes are binary, a cross-entropy loss is appropriate, whereas for continuous potential outcomes, an MSE loss is preferable.

In summary, in the first-stage training, we use the following training objective:
\begin{equation*}
\min_{\phi,\psi,h_0,h_1}\mathcal{L}_f+\lambda_1 \mathcal{L}_{rec}+\lambda_2 \mathcal{L}_{unb},
\end{equation*}
where $\lambda_1>0$ and $ \lambda_2>0$ are tunable hyperparameters.
\subsection{Second Stage: Finetuning Stage}
In the second stage of training, we exploit the small-scale unconfounded RCT data to remove the hidden confounding by concatenating the learned covariate representation in the first stage with a newly learned augmented covariate representation, then finetuning the prediction heads to obtain an unbiased CATE estimation. To achieve this, we keep the biased representation $Z$ produced by $\phi$ unchanged but only treat it as a part of the representation, together with an additional $Z^{U}$ generated by another representation module $\phi^U$. We call $\phi^U$ a representation adapter, as it helps to adapt the final representation to account for the hidden confounding in the observed data. In addition, a large proportion of the parameters of the prediction heads $g_0$ and $g_1$ are initialized by $h_0$ and $h_1$, respectively. With the above steps, the second stage aims at adjusting the hidden confounding through the augmented covariate representation and the finetuned prediction heads with partial parameter initialization. Below we explain the modules in detail.

\subsubsection{Representation Adapter}
We employ a shallower feed-forward network $\phi^U$ as the representation adapter. It is worth noting that the width and depth of $\phi^U$ can be adjusted based on the scale of RCT data size. If the size of RCT data was comparable to that of OBS data, we can use the same architecture as $\phi$. However, in real-world cases, the RCT data is rare compared to the OBS data, so the size of $\phi^U$ should be smaller. We denote the representation generated by $\phi^U$ as $Z^U$. To make sure that $Z^U$ captures different features of covariates from $Z$, we employ mutual information to control the overlap between the two covariate representations:
\begin{equation*}
\mathcal{L}_{MI}=CLUB(Z,Z^U),
\end{equation*}
\begin{equation*}
\begin{aligned}
 CLUB(Z,Z^U)&=\frac{1}{m^2} \sum_{i\in \mathcal{R}} \sum_{j\in \mathcal{R}}\Big[\log q_\theta(Z^U_i \mid Z_i)-\log q_\theta(Z^U_j \mid Z_i)\Big] \\
& =\frac{1}{m} \sum_{i\in \mathcal{R}}\Big[\log q_\theta(Z^U_i \mid Z_i)-\frac{1}{m} \sum_{j\in\mathcal{R}} \log q_\theta(Z^U_j \mid Z_i)\Big],
\end{aligned}
\end{equation*}
where $CLUB(Z,Z^U)$ is the empirical Contrastive Log-ratio Upper Bound (CLUB) of mutual information~\cite{cheng2020club} between two covariate representations $Z$ and $Z^U$,
and $q_\theta (Z^U\mid Z)$ is the variational approximation of $P(Z^U\mid Z)$. With good variational approximation $q_\theta (Z^U\mid Z)$, it can be shown that the empirical CLUB is still a valid upper bound of the ground-truth mutual information.
\subsubsection{Prediction}
Similarly to the first stage, we design two $l_p-$layer prediction heads $g_0$ and $g_1$ to estimate the potential outcomes under control and treatment groups with hidden confoundings, respectively. Notice that $g_0,g_1$ and $h_0,h_1$ have the same depth $l_p$, yet every layer of $g_t$ has a larger or equal width than $h_t$. For $t\in\{0,1\}$, we define the header $h_t$ as:
\begin{equation*}
\begin{aligned}
    & a_{h_t}^{(0)}=Z, ~a_{h_t}^{(l)}=\sigma(W_{h_t}^{(l)}a_{h_t}^{(l-1)}+b_{h_t}^{(l)}), \text{ for } l=1,2,\ldots,l_p-1,\\
    &a_{h_t}^{(l_p)}=\tilde{Y}_t=W_{h_t}^{(l_p)}a_{h_t}^{(l_p-1)}+b_{h_t}^{(l_p)},
\end{aligned}
\end{equation*}
where $W_{h_t}^{(l)}$ is the weight matrix from layer $l-1$ to layer $l$, $b_{h_t}^{(l)}$ is the bias vector of layer $l$, $a_{h_t}^{(l)}$ is the output of layer $l$ for $l\in \{1,2,\ldots,l_p\}$ and $\sigma$ is the activation function. The definition of $g_t$ is:
\begin{equation*}
\begin{aligned}
    & a_{g_t}^{(0)}=\begin{bmatrix}
Z \\
Z^U
\end{bmatrix},~a_{g_t}^{(l)}=\sigma(W_{g_t}^{(l)}a_{g_t}^{(l-1)}+b_{g_t}^{(l)}), \text{ for } l=1,2,\ldots,l_p-1,\\
    & a_{g_t}^{(l_p)}=\hat{Y}_t=W_{g_t}^{(l_p)}a_{g_t}^{(l_p-1)}+b_{g_t}^{(l_p)},
\end{aligned}
\end{equation*}
with $W_{g_t}^{(l)}, b_{g_t}^{(l)}, a_{g_t}^{(l)}$ having similar meanings as $W_{h_t}^{(l)}, b_{h_t}^{(l)}, a_{h_t}^{(l)}$ but for $g_t$, $\hat{Y}_t$ is the final prediction for potential outcome under treatment $t$. In our design, every layer of $g_t$ has a larger or equal width than $h_t$, thus the dimension of $W_{g_t}^{(l)}, b_{g_t}^{(l)}, a_{g_t}^{(l)}$ is no less than that of $W_{h_t}^{(l)}, b_{h_t}^{(l)}, a_{h_t}^{(l)}$, respectively. We divide the parameters of $g_t$ as:
\begin{equation*}
W_{g_t}^{(l)}=\begin{bmatrix}
    ^{1}W_{g_t}^{(l)},^{2}W_{g_t}^{(l)}\\
^{3}W_{g_t}^{(l)}, ^{4}W_{g_t}^{(l)}
\end{bmatrix}, b_{g_t}^{(l)}=\begin{bmatrix}
^{1}b_{g_t}^{(l)} \\
^{2}b_{g_t}^{(l)}
\end{bmatrix},
\end{equation*}
where $^{1}W_{g_t}^{(l)}, ^{1}b_{g_t}^{(l)}$ have the same shapes as $W_{h_t}^{(l)}, b_{h_t}^{(l)}$ respectively, for $l=1,2,\ldots,l_p$, with the detailed initialization strategy as follows.

\paragraph{Initialization.}
The initialization of the model parameters is crucial for the preservation of covariate information from the first stage as well as the effectiveness of the finetuning stage. The goal of initialization is to make sure the model initially produces the same prediction as the trained first-stage model. Nonetheless, a challenge of parameter initialization is that the model architecture of the second stage differs from that of the first stage, because of the augmented covariate representation. Based on the division of the parameters, we propose the following initialization strategy:
\begin{equation*}
(^{1}W_{g_t}^{(l)},^{2}W_{g_t}^{(l)},^{3}W_{g_t}^{(l)},^{4}W_{g_t}^{(l)},^{1}b_{g_t}^{(l)},^{2}b_{g_t}^{(l)}) \leftarrow (W_{h_t}^{(l)},0,0,0,b_{h_t}^{(l)},0)    
\end{equation*}
for $l=1,2,\ldots,l_p$. That is, the shared parameters between the prediction heads $g_t$ and $h_t$ are initialized to be the same, and the rest parameters of the prediction head $g_t$ are initialized to be zero.

As in the first stage, we denote $f_g(x,t)=g_t ([\phi(x)^\intercal|\phi^U(x)^\intercal]^\intercal)$ for $t\in \{0,1\}$, where $[\phi(x)^\intercal|\phi^U(x)^\intercal]$ is the concatenated covariate representation. Given the RCT data $\{(X_i,T_i,Y_i,G_i=1): i\in \mathcal{R}\}$, the prediction loss is computed as:
\begin{equation*}
 \mathcal{L}_{pred}=\frac{1}{|\mathcal{R}|}\sum_{i\in \mathcal{R}}w_i\cdot l(Y_{i},f_g(X_i, T_i)),
\end{equation*}similarly with $w_i=\frac{T_i}{2u}+\frac{1-T_i}{2(1-u)}$ and $u=\frac{1}{m}\sum_{i=n+1}^{n+m}T_i$.

\paragraph{Regularization.}
Since small-scale RCT data may cause overfitting during the finetuning phase, we further introduce a constraint to ensure that the parameters of the second-stage finetuned model do not deviate significantly from the parameters of the first-stage trained model. Denote the initial value of $\theta_{g_t}$ as $\theta_{g_t}^0$, to constrain the deviation from the initial value, we propose to use $l_2$-norms in the training loss objective:
\begin{equation*}
\mathcal{L}_{shift}=||\theta_{g_0}-\theta_{g_0}^0||_2^2+||\theta_{g_1}-\theta_{g_1}^0||_2^2.
\end{equation*}
Overall, the training loss of the second stage is given by:
\begin{equation*}
\min_{\phi^U,~g_0,~g_1}\mathcal{L}_{pred}+\lambda_3 \mathcal{L}_{MI}+\lambda_4 \mathcal{L}_{shift},
\end{equation*}
where $\lambda_3$ and $\lambda_4$ are tunable hyperparameters. Note that during the second-phase training, we froze the parameters of the representation module $\phi$ and the decoder network $\psi$, while train the representation adapter module $\phi^U$ and the two prediction heads $g_0, g_1$. We summarize the whole learning algorithm in Alg.~\ref{alg.1}.
\begin{algorithm}[t]
\caption{Learning algorithm of the TSPF framework.}
\label{alg.1}
\KwIn{OBS data $\mathcal{D}^{OBS}=\{(X_i,T_i,Y_i,G_i=0)\}_{i=1}^n$, RCT data $\mathcal{D}^{RCT}=\{(X_i,T_i,Y_i,G_i=1)\}_{i=n+1}^{n+m}$ and four hyperparameters $\lambda_k>0, k=1,\ldots,4.$}
Compute $w_i=\frac{T_i}{2u}+\frac{1-T_i}{2(1-u)}$ with $u=\frac{1}{n}\sum_{i=1}^{n}T_i$ for $i=1,...,n$\;
\For{number of steps for training the first-stage model}{Sample a batch $\{(X_i,T_i,Y_i)\}_{i\in B}$ from $\mathcal{D}^{OBS}$\;
Update $\theta_1=(\theta_\phi,\theta_\psi,\theta_{h_0}, \theta_{h_1})$ by descending along the gradient $\nabla_{\theta_1} (\mathcal{L}_f+\lambda_1 \mathcal{L}_{rec}+\lambda_2 \mathcal{L}_{unb})$\;} 
Initialize $(^{1}W_{g_t}^{(l)},^{2}W_{g_t}^{(l)},^{3}W_{g_t}^{(l)},^{4}W_{g_t}^{(l)},^{1}b_{g_t}^{(l)},^{2}b_{g_t}^{(l)})$ by $ (W_{h_t}^{(l)},0,0,0,b_{h_t}^{(l)},0)$ for $l=1,2,\ldots,l_p$ and $t=0,1$\;
Compute $w_i=\frac{T_i}{2u}+\frac{1-T_i}{2(1-u)}$ with $u=\frac{1}{m}\sum_{i=n+1}^{n+m}T_i$ for $i=n+1,...,n+m$\;
\For{number of steps for training the second-stage model}{Sample a batch $\{(X_i,T_i,Y_i)\}_{i\in B}$ from $\mathcal{D}^{RCT}$\;
Update $\theta_2=(\theta_{\phi^U}, \theta_{g_0}, \theta_{g_1})$ by descending along the gradient $\nabla_{\theta_2} (\mathcal{L}_{pred}+\lambda_3 \mathcal{L}_{MI}+\lambda_4 \mathcal{L}_{shift})$\;} 
\end{algorithm}

Compared to residual correction methods as in Kallus et al.~\cite{kallus2018removing}, our representation adapter module guarantees a stronger representation ability, relaxing the linearly additive assumption. More importantly, when RCT data are limited, our proposed partial initialization strategy in the TSPF framework can avoid overfitting.

\begin{table*}[]
\centering
\caption{The experiment results on the \textsc{IHDP} dataset and \textsc{Jobs} dataset. The best result is bolded and the second best is underlined.}
\resizebox{1\linewidth}{!}{
\begin{tabular}{l|llll|llll}
\toprule
          & \multicolumn{4}{c|}{\textbf{IHDP}}                                                                                & \multicolumn{4}{c}{\textbf{Jobs}}                                                                                \\ \cmidrule{2-9} 
          & \multicolumn{2}{c}{In-sample}                  & \multicolumn{2}{c|}{Out-sample}                  & \multicolumn{2}{c}{In-sample}                  & \multicolumn{2}{c}{Out-sample}                  \\ \midrule
Methods   & \multicolumn{1}{c}{$\sqrt{\epsilon_{\text{PEHE}}}$} & \multicolumn{1}{c}{$\epsilon_{\text{ATE}}$} & \multicolumn{1}{c}{$\sqrt{\epsilon_{\text{PEHE}}}$} & \multicolumn{1}{c|}{$\epsilon_{\text{ATE}}$} & \multicolumn{1}{c}{$\sqrt{\epsilon_{\text{PEHE}}}$} & \multicolumn{1}{c}{$\epsilon_{\text{ATE}}$} & \multicolumn{1}{c}{$\sqrt{\epsilon_{\text{PEHE}}}$} & \multicolumn{1}{c}{$\epsilon_{\text{ATE}}$} \\
\cmidrule{1-9}
T-learner & 0.44 $\pm$ 0.03                & 0.04 $\pm$ 0.02               & 0.52 $\pm$ 0.05                & \textbf{0.02 $\pm$ 0.01}                & 0.66 $\pm$ 0.27                         & \underline{0.02 $\pm$ 0.02}                        &  0.60 $\pm$ 0.20                        & \underline{0.02 $\pm$ 0.01}                        \\
S-learner & 0.98 $\pm$ 0.18                & 0.04 $\pm$ 0.03               & 1.37 $\pm$ 0.34                & 0.15 $\pm$ 0.10                & 0.70 $\pm$ 0.30                         & \underline{0.02 $\pm$ 0.02}                        &  0.67 $\pm$ 0.39                        & 0.03 $\pm$ 0.02                       \\
DR-learner & 0.71 $\pm$ 0.19                & 0.06 $\pm$ 0.04               & 0.81 $\pm$ 0.25                & 0.06 $\pm$ 0.03                & 0.50 $\pm$ 0.07                         & 0.07 $\pm$ 0.02                        & 0.48 $\pm$ 0.09                         & 0.07 $\pm$ 0.03                        \\
SCIGAN  & 2.53 $\pm$ 0.47                         & 0.63 $\pm$ 0.29                        & 2.58 $\pm$ 0.57                        & 0.55 $\pm$ 0.48  & 2.28 $\pm$ 0.75                & 0.56 $\pm$ 0.15               & 2.15 $\pm$ 0.80                & 0.47 $\pm$ 0.21                                        \\
Causal Forest & 1.90 $\pm$ 0.29                & 0.07 $\pm$ 0.05               & 2.04 $\pm$ 0.42                & 0.19 $\pm$ 0.11                & 1.38 $\pm$ 0.40                         & 0.13 $\pm$ 0.08                        & 1.23 $\pm$ 0.35                         & 0.14 $\pm$ 0.08                        \\
TARNet  & 0.42 $\pm$ 0.09                         & 0.05 $\pm$ 0.04                        & 0.44 $\pm$ 0.12                         & 0.05 $\pm$ 0.05   & 0.19 $\pm$ 0.17                & \underline{0.02 $\pm$ 0.01}               & 0.13 $\pm$ 0.02                & \underline{0.02 $\pm$ 0.01}                                       \\
DragonNet & \underline{0.19 $\pm$ 0.04}                & \underline{0.03 $\pm$ 0.02}               & \underline{0.26 $\pm$ 0.08}                & \underline{0.04 $\pm$ 0.02}                & \underline{0.15 $\pm$ 0.11}                         & \textbf{0.01 $\pm$ 0.01}                        & \underline{0.11 $\pm$ 0.03}                         &  \textbf{0.01 $\pm$ 0.01}                       \\
DESCN     & 0.28 $\pm$ 0.06                & 0.05 $\pm$ 0.05               & 0.41 $\pm$ 0.11                & 0.07 $\pm$ 0.06                & 0.44 $\pm$ 0.09                         & 0.28 $\pm$ 0.13                        & 0.44 $\pm$ 0.08                         & 0.27 $\pm$ 0.13                        \\
DRCFR     & 0.74 $\pm$ 0.32                & 0.15 $\pm$ 0.10               & 0.90 $\pm$ 0.52                & 0.18 $\pm$ 0.17                & 0.91 $\pm$ 0.49                         & 0.08 $\pm$ 0.08                        & 0.71 $\pm$ 0.43                         & 0.09 $\pm$ 0.07                        \\

CorNet  & 0.74 $\pm$ 0.32                & 0.15 $\pm$ 0.10               & 0.90 $\pm$ 0.52                & 0.18 $\pm$ 0.17                & 0.91 $\pm$ 0.49                         & 0.08 $\pm$ 0.08                        & 0.71 $\pm$ 0.43                         & 0.09 $\pm$ 0.07                        \\
TSFP (ours)      & \textbf{0.13 $\pm$ 0.02}                & \textbf{0.02 $\pm$ 0.02}               & \textbf{0.16 $\pm$ 0.04}                & \underline{0.04 $\pm$ 0.02}                &  \textbf{0.09 $\pm$ 0.03}                          & \textbf{{0.01 $\pm$ 0.01} }                                             & \textbf{0.06 $\pm$ 0.01}                         & \textbf{0.01 $\pm$ 0.01}                     \\ \bottomrule
\end{tabular}}
\vspace{-6pt}
\label{tab:real}
\end{table*}
\section{Experiment}
\subsection{Datasets} Following previous studies~\cite{shalit2017estimating,louizos2017causal,yoon2018ganite}, we conduct experiments on two semi-synthetic datasets, namely \textbf{IHDP}~\cite{hill2011bayesian} and \textbf{Jobs}~\cite{shalit2017estimating}. The \textbf{IHDP} is a semi-synthetic dataset for causal effect estimation. The dataset is based on the Infant Health and Development Program, where the covariates are obtained by a randomized experiment investigating the
effect of home visits by specialists on future cognitive scores. It consists of 747 units (19\% treated, 81\% control) and 25 covariates measuring the children and
their mothers. The \textbf{Jobs} is a common benchmark dataset developed
by LaLonde in 1986, studying the change of income and employment status after job training. We use an extended version of \textbf{Jobs} that comprises about 3,000 units (10\% treated, 90\%
control) with 17 covariates.

\noindent
\subsection{Data Preprocessing}

For both \textbf{IHDP} and \textbf{Jobs}, we simulate hidden confounding by generating a $c$-dimensional confounder $U_i\in \mathbb{R}^c$. To make sure the $U_i$ has a non-zero effect on $Y_i$ and $T_i$, we generate the data below:
\begin{equation*}
    \begin{aligned}
        &W_1\sim \mathcal{N}(0,0.1)^d,~~~ W_2\sim\mathcal{N}(0.02,0.1)^c,~~~ W_3\sim\mathcal{N}(0.1,1)^d,\\
        &W_4\sim\mathcal{N}(0.1,1)^c,~~~ W_5\sim\mathcal{U}(0,0.2)^d,~~~W_6\sim\mathcal{U}(0,0.2)^c,\\
        &U_i\sim \mathcal{U}(0,0.2)^c,~~~ T_i \sim \text{Bern}(\sigma(W_1\cdot X_i+W_2\cdot U_i)),\\
        &\mu_i^0=W_3\cdot X_i+W_4\cdot U_i,~~~ \mu_i^1=W_5\cdot X_i+W_6\cdot U_i+4,\\
        &Y_i^0\sim\mathcal{N}(\mu_i^0,0.1),~~~ Y_i^1\sim\mathcal{N}(\mu_i^1,0.1),
    \end{aligned}
\end{equation*}where $\mathcal{N}(\mu, D)$ denotes the normal distribution with mean $\mu$ and variance $D$, $\mathcal{U}(a,b)$ is the uniform distribution on interval $(a,b)$, $\text{Bern}(p)$ means the Bernoulli distribution with probability $p$, $\sigma(x)=1/(1+\exp(-x))$ is the sigmoid function. Note that we keep $E[Y_i^1] = E[Y_i^0] + 4$ as the same as the \textbf{IHDP} dataset and we let $E[W_i] > 0, i \in \{2, 4, 6\}$ to ensure the non-zero effect of $U_i$. The hidden confounding strength parameter $c$ is set to 30 in our experiments. Then we slice the training, validation, and test sets in the ratio of 63/27/10. In addition, to obtain a separate RCT training dataset for data fusion,  we first randomly split 10\% of the training samples, and then assign treatments $T^{new}_i$ according to the following formula and replace the factual treatment $T_i$ and outcome $Y^f_i$ to obtain a RCT dataset:
\begin{equation*}
    T^{new}_i = \operatorname{Bern}(0.5), ~~~Y^{new}_i = \mathbb{I}\{T^{new}_i = T_i\}(Y^f_i - Y^{cf}_i) + Y^{cf}_i,
\end{equation*}
where $\mathbb{I}$ is the indicator function, $Y^f_i=T_iY_i^1+(1-T_i)Y_i^0$ is the factual outcome, and $Y^{cf}_i=T_iY_i^0+(1-T_i)Y_i^1$ is the counterfactual outcome. Finally, we replace the treatment $T_i$ and factual outcome $Y_i$ using the above formula for all samples in the validation set.

\noindent
\subsection{Baselines and Evaluation Metrics}

\subsubsection{Baselines}
\begin{itemize}
  \item T-learner~\cite{kunzel2019metalearners}: T-learner utilizes two separate regressors for each treatment group.
  \item S-learner~\cite{Athey2015MachineLM}: S-learner treats the indicator of treatment as features, utilizing a single model to estimate the potential outcome for both treatment and control groups.
  \item DR-learner~\cite{DR-learner}: DR-learner estimates the CATE via cross-fitting a doubly robust score function in two stages.
      \item SCIGAN~\cite{yoon2018ganite}: SCIGAN utilizes a generative adversarial network to model treatment effect.
    \item Causal Forest~\cite{wager2018estimation}: Causal Forest is a random forest-based model that directly estimates the treatment effect.      
    \item TARNet~\cite{shalit2017estimating}: TARNet applies a shared representation layer and a two-head network inference layer.

    \item DragonNet~\cite{shi2019adapting}: DragonNet designs an adaptive neural network to learn propensities and counterfactual outcomes.
    \item DESCN~\cite{zhong2022descn}: DESCN uses deep networks to model treatment effects in the entire sample space.
    \item DRCFR~\cite{hassanpour2019learning}: DRCFR aims to learn disentangled representations of covariates and address selection bias in CATE estimation.
    \item CorNet~\cite{hatt2022combining}: CorNet leverages OBS data to learn a biased estimate for the treatment effect and then aims to estimate the non-linear bias using RCT data.
\end{itemize}
\subsubsection{Evaluation Metrics}
Following previous studies~\cite{shalit2017estimating, yao2018representation}, we evaluate the performance of CATE estimation using \emph{the square root of Precision in Estimation of Heterogeneous Effects} (PEHE): 
\begin{equation*}
    \sqrt{{\epsilon_{\text{PEHE}}}} =
\sqrt{{\frac{1}{n} \sum_{i=1}^n ((\hat {Y}_{i}^1 - \hat {Y}_{i}^0) - (Y_{i}^1 - Y_{i}^0))^2}},
\end{equation*}
where $\hat {Y}_{i}^t$ and $Y_{i}^t$ are the predicted and ground truth values for the potential outcomes of individual $i$ under treatment $t$. In addition, we also use the \emph{absolute error in Average Treatment Effect} (ATE) to evaluate estimation performance, which is defined as: \begin{equation*}
    \epsilon_{\text{ATE}}=\frac{1}{n}\left| \sum_{i=1}^n ((\hat {Y}_{i}^1 - \hat {Y}_{i}^0) - (Y_{i}^1 - Y_{i}^0))\right|.
\end{equation*}

\noindent

\subsubsection{Experimental Details}
We adopt a multi-layer perceptron~\cite{haykin1994neural} with 2 layers for our representation and reconstruction modules as well as the prediction heads in both stages. We tune the scale parameters in the loss functions from $1e-5$ to $0.1$. For the baselines with only one stage, we combine the OBS and RCT samples as the training dataset. We report both in-sample and out-of-sample results for $\sqrt{\epsilon_{\mathrm{PEHE}}}$ and $\epsilon_{\mathrm{ATE}}$ metrics in our experiments.
\subsection{Performance Analysis}
Table \ref{tab:real} shows the prediction performance with varying baselines and our methods. First, representation-based methods generally outperform generation-based methods and meta-learners, which shows the effectiveness of causal representation learning. Note that our TSPF exhibits the most competitive performance in most cases, outperforming both the representation-based and generation-based methods. More importantly, TSPF significantly outperforms another two-stage method CorNet for all metrics on the two datasets.
\section{Conclusion}
This paper studies the CATE estimation problem in the presence of hidden confounding for fusing large-scale OBS data and small-scale RCT data. We propose a two-stage pretraining-finetuning framework to tackle the overfitting problem caused by the small-scale RCT data. Specifically, the foundational representation learned in the first stage is used to adjust for the \emph{measured} confounding bias in the OBS data. The augmented representation learned in the second stage is used to mitigate for the \emph{hidden} confounding bias guided by the RCT data. To avoid overfitting caused by the small-scale RCT data in the second stage, instead of training a separate network, we propose to partially initialize the network parameters from the pretrained network from the first stage. Compared to the previous CATE estimation methods that combine OBS and RCT data, our approach has the advantage of not restricting the data-generating process (e.g., linearity or additive noise assumptions) and suffering from overfitting. The experiments conducted in the real-world datasets demonstrate the superiority of our approach.



\begin{acks}
HL was supported by National Natural Science Foundation of China (623B2002). MG was supported by ARC DE210101624 and ARC DP240102088.
\end{acks}

\bibliographystyle{ACM-Reference-Format}
\bibliography{sample-base}










\end{document}